\documentclass[conference]{IEEEtran}

\usepackage{cite}
\usepackage{amsmath,amssymb,amsfonts}
\usepackage{booktabs}
\usepackage{graphicx}
\usepackage{textcomp}
\usepackage[hyphens]{url}

\graphicspath{{figures/}}

\def\BibTeX{{\rm B\kern-.05em{\sc i\kern-.025em b}\kern-.08em
    T\kern-.1667em\lower.7ex\hbox{E}\kern-.125emX}}

\newif\ifwithappendices%
\ifdefined\WITHAPPENDICES%
    \withappendicestrue%
\else
    \withappendicesfalse%
\fi

\begin{document}

\title{BAC-JEPA\@: Label-Efficient Breast Arterial Calcification Segmentation via Synthetic Mammography-Guided Supervision}

\author{\IEEEauthorblockN{Scott Chase Waggener and Lakshman Tamil}
\IEEEauthorblockA{\textit{Department of Computer Engineering} \\
\textit{University of Texas at Dallas}\\
Richardson, TX \\
scott.waggener@utdallas.edu; laxman@utdallas.edu}
}

\maketitle

\begin{abstract}
Breast arterial calcification (BAC) visible on screening mammograms is an emerging
imaging biomarker for cardiovascular risk, but quantitative use requires
reproducible segmentation and expert pixel-level annotations are expensive. We
present BAC-JEPA, a label-efficient segmentation framework that trains on
procedurally generated arterial calcification inserted into real mammographic
backgrounds with exact masks. Candidate backgrounds were selected from
model-screened mammograms with low predicted BAC response; the generator
samples arterial graphs, disease burden, radiographic appearance, and
hard-negative distractors, including nonarterial calcifications and metallic
objects. Synthetic masks are paired with mammography self-supervised Vision
Transformer encoders and a high-resolution convolutional decoder to produce
full-resolution segmentation maps. The study used 75,472 mammography studies
from 34,956 patients for background selection and representation learning,
trained on synthetic images from 10,000 training backgrounds, selected
checkpoints with 1,000 development backgrounds, and evaluated transfer on all
1,000 human-labeled BacSeg synthetic two-dimensional mammograms derived from
digital breast tomosynthesis (synthetic 2D mammograms). On held-out
synthetic validation data, the larger backbone achieved intersection-over-union
0.5325 and Dice 0.6357. On BacSeg, image-level classification from segmentation
probability maps reached an area under the receiver operating characteristic
curve of 0.8719, with 0.8547 for the smaller backbone. Four-view
inference required 110.68--213.63 ms on an RTX 5090 graphics processor, and
synthetic image generation at the most severe preset averaged 2.7071 s per
image on a multicore workstation. These results indicate that breast arterial
calcification-specific synthetic supervision can produce useful image-level
transfer without human pixel-level training masks, while expert-reviewed
real-mammogram segmentation remains necessary for clinical validation and
calibration.
\end{abstract}

\begin{IEEEkeywords}
Breast arterial calcification, mammography, medical image segmentation,
synthetic data, self-supervised learning, Vision Transformers.
\end{IEEEkeywords}

\section{Introduction}
Cardiovascular disease remains a leading cause of morbidity and mortality
among women, yet conventional risk assessment commonly relies on clinical and
laboratory variables rather than direct vascular assessment~\cite{Dapamede2026}.
Screening mammography creates an opportunity for cardiovascular risk assessment
because it is already widely performed in asymptomatic women and can reveal
breast arterial calcification (BAC), a benign finding unrelated to breast cancer
but increasingly linked to vascular aging and cardiovascular risk.
Professional groups have argued that BAC reporting may support cardiovascular
risk communication using information already visible on routine mammograms; the
Canadian Society of Breast Imaging notes that BAC is seen in
approximately 12--42.5\% of screening mammograms and is not routinely reported
in many settings~\cite{CSBI2023}.

Recent cohort studies strengthen the rationale for quantitative BAC assessment.
In the MINERVA cohort, BAC was associated with elevated risk of cardiovascular
events among postmenopausal women~\cite{Iribarren2022}. Allen et al.\ studied
18,092 women and found that an automated BAC score was associated with
cardiovascular outcomes and all-cause mortality after adjustment for clinical
risk factors~\cite{Allen2024}. Dapamede et al.\ extended this evidence to
123,762 women across internal and external health-system cohorts, using an
artificial-intelligence model to quantify BAC and showing dose-response
associations with major adverse cardiovascular events and
mortality~\cite{Dapamede2026}.

The clinical utility of BAC depends not only on detecting its presence but also
on measuring its extent reproducibly. Binary detection can support
opportunistic reporting, whereas segmentation enables quantitative biomarkers
such as BAC area, length, burden, and severity category for longitudinal
assessment and risk stratification. Prior BAC systems therefore treat
segmentation as a central step in automated mammography-based cardiovascular
assessment, with automated measurements showing agreement with expert
annotations in several settings~\cite{Guo2021,Dapamede2026}.

BAC segmentation is technically difficult. BAC appears as thin, elongated,
high-intensity arterial tracks, sometimes with parallel tram-track morphology,
against heterogeneous breast parenchyma. The target occupies a small fraction
of the image, creating severe class imbalance, and can be confused with
nonvascular calcifications, vessel edges, dense tissue, skin artifacts, and
other high-contrast structures. Dense pixel-level annotation is expensive and
requires domain expertise. Wang et al.\ showed that annotation quality can
materially affect BAC segmentation: after identifying and correcting false
positive labels using morphology- and physics-informed rules, retraining
improved Dice performance by 29\% relative to evaluation on corrected
labels~\cite{Wang2022}.

These challenges motivate label-efficient approaches that reduce dependence on
large manually segmented BAC datasets while preserving validity on real
mammograms. Synthetic data offers one route: realistic BAC patterns can be
inserted into real mammographic backgrounds with known geometry, producing
paired image-mask examples at low annotation cost. However, most synthetic
mammography work has focused on whole-image simulation, masses,
microcalcification clusters, or generic abnormality synthesis rather than BAC
morphology~\cite{Badano2018,Sharma2019,Sizikova2023,Oyelade2022,MontoyaDelAngel2024}.
BAC is not a compact mass or clustered lesion; it is vessel-associated,
elongated, often fragmented, and clinically measured through morphology,
topology, and burden. A useful BAC generator must therefore model arterial
geometry, local intensity, blur, partial-volume effects, and plausible
anatomical placement.

We propose a synthetic BAC generation framework for training segmentation
models on real mammographic backgrounds. The generator inserts anatomically and
radiographically plausible BAC patterns while producing exact pixel-level
masks. We pair this supervision with a Vision Transformer segmentation model
initialized using Joint Embedding Predictive Architecture (JEPA)-style
self-supervised pre-training, combining
mammography representations learned from unlabeled images with scalable exact
masks for dense BAC supervision~\cite{Dosovitskiy2021,Assran2023}.

Our study evaluates whether synthetic supervision can produce useful BAC
segmentation behavior without human pixel-level BAC masks for training. We
report held-out synthetic segmentation metrics against exact masks, image-level
external validation on the public human-labeled BacSeg dataset, and runtime
benchmarks for inference and image generation. The primary contribution is a
practical framework for generating BAC-specific supervision and testing
transfer to independent human-labeled synthetic two-dimensional mammograms
derived from digital breast tomosynthesis (synthetic 2D mammograms).

\emph{Contributions.} This paper makes four contributions: (1) a BAC-specific
synthetic generation framework that inserts realistic arterial calcification
patterns into real mammograms and produces exact pixel-level segmentation
masks; (2) a label-efficient training strategy combining JEPA-pretrained Vision
Transformer representations with synthetic BAC supervision; (3) evaluation of
synthetic pixel-overlap performance together with independent image-level area
under the receiver operating characteristic curve (AUROC) on human-labeled
BacSeg synthetic 2D mammograms; and (4) benchmarks
quantifying four-view inference cost and end-to-end synthetic BAC generation
throughput.

\section{Related Work}

\subsection{BAC as a Cardiovascular Imaging Biomarker}

BAC has historically been treated as an incidental benign mammographic finding,
but recent evidence reframes it as a sex-specific cardiovascular imaging
biomarker associated with vascular aging and adverse cardiovascular
outcomes~\cite{Iribarren2022,Dapamede2026}. Because screening mammography is
already widely performed, professional groups have argued that BAC reporting
could add cardiovascular risk information without a separate imaging
examination~\cite{CSBI2023}. This motivates quantitative segmentation rather
than presence-only detection: risk modeling and longitudinal monitoring benefit
from reproducible estimates of BAC area, length, burden, and severity.

\subsection{Automated BAC Segmentation and Label Efficiency}

Deep learning methods for BAC analysis commonly use dense segmentation followed
by burden estimation. DU-Net applied a convolutional network to arterial
calcification detection~\cite{AlGhamdi2020}, SCU-Net introduced a fine-vessel
segmentation architecture with automated measurement agreement~\cite{Guo2021},
and later systems explored label correction, recurrent attention,
generative-adversarial formulations, and U-Net variants for BAC
quantification~\cite{Wang2022,Alamir2023,AlJabri2024,Li2025}. These studies
show feasibility, but limited and variable expert masks remain a bottleneck:
Wang et al.\ showed that false-positive labels can substantially degrade
training and that morphology-informed correction improves segmentation
performance~\cite{Wang2022}. Our approach targets this bottleneck from a
complementary direction by using synthetic BAC insertion to provide exact masks
with controlled burden, morphology, prevalence, and difficulty.

\subsection{Synthetic Mammography and Transformer Pre-Training}

Synthetic data has a substantial history in mammography, including virtual
imaging trials, large synthetic mammography resources, and generative models
for mammographic abnormalities~\cite{Badano2018,Sizikova2023,Oyelade2022,MontoyaDelAngel2024}.
However, calcification insertion work has primarily targeted compact
microcalcification findings rather than elongated vascular BAC\@; for example,
MC-GenRef generates exact image-mask pairs for microcalcification segmentation
by injecting plausible patterns into real negative patches~\cite{Cho2026}. The
different clinical meaning and vessel-associated topology of BAC motivate
BAC-specific synthesis rather than reuse of generic lesion or
microcalcification generators. In parallel, Vision Transformers and image-based
Joint-Embedding Predictive Architecture (I-JEPA) provide a natural modeling
context: JEPA-style pre-training can learn
mammographic anatomy and imaging statistics from unlabeled data, while
synthetic BAC generation supplies exact dense supervision~\cite{Dosovitskiy2021,Assran2023}.

\section{Methods}

\subsection{Data}

The study dataset combines public and licensed mammography data, including
CBIS-DDSM film mammograms from the United States~\cite{Lee2017CBISDDSM}, the
OPTIMAM Mammography Image Database (OMI-DB) from United Kingdom breast
screening centers~\cite{HallingBrown2021}, and MedCognetics licensed and
proprietary mammograms collected from multiple clinical sites. The dataset
contains 75,472 studies from 34,956 patients and 338,537 images. It is
partitioned at the study level into training, development, and held-out test
sets. The training, development, and held-out internal test partitions came from
independent clinical sites, so no clinical site contributed images to more than
one partition. The training partition contains 71,103 studies from 31,110
patients and 307,932 images, the development set contains 2,107 studies from
1,584 patients and 12,509 images, and the held-out test set contains 2,262
studies from 2,262 patients and 18,096 images. The site-disjoint test partition
was reserved for later evaluation and was not used to select synthetic training
backgrounds.

Independent human-labeled evaluation used the BacSeg dataset reported by
AlJabri et al.~\cite{AlJabri2024}, a public segmentation dataset of
craniocaudal (CC) and mediolateral-oblique (MLO) synthetic 2D mammograms
with BAC ground-truth masks produced under expert breast-radiologist
supervision. The publicly
available download used in this study contained 1,000 images across the provided
train, validation, and test splits, and we evaluated all available images. This
evaluation set is independent of our generator and model-screened background
pool.

The non-BacSeg study corpus used for training, development, and internal testing
did not include native image-level or pixel-level BAC annotations. We therefore
scored mammograms with SCU-Net, a previously published BAC segmentation
model~\cite{Guo2021}, and restricted candidate BAC-free background selection to
training and development images with complete partition metadata. This eligible
pool contained 207,615 training images and 8,184 development images. Standard
2D craniocaudal (CC) and mediolateral-oblique (MLO) mammograms were assigned a
global, threshold-free BAC score
summarizing BAC-like model response over the breast foreground, then ranked
within each partition. We selected the 10,000 lowest-scoring training images and
1,000 lowest-scoring development images as candidate BAC-free backgrounds for
synthetic insertion. These images are treated as BAC-free for generation, but the
designation is model-screened rather than expert-confirmed.

\subsection{Synthetic BAC Generation}\label{subsec:synthetic-bac-generation}

Synthetic training examples were produced by inserting procedurally generated
BAC into proxy BAC-free mammograms while retaining exact pixel labels by
construction. Each selected background was cropped, resized, and padded to
2048~$\times$~1536 after estimating breast foreground, pectoral muscle, border
artifacts, and exclusion margins that define a safe insertion region.

Within this region, the generator samples a projected arterial graph in
physical coordinates, guided by multi-scale curvilinear image structure so that
inserted BAC follows plausible background anatomy. Severity presets control
the number of involved vessels, calcified length, run continuity, gap
frequency, and radiopacity. Calcified runs are rendered as center-cut,
vessel-constrained microcalcification-like, or mixed deposits with granular
texture, porosity, local width variation, and faint trace-BAC vessel-wall
opacity to approximate single-wall, tram-track, and subtle vessel-associated
appearances.

The rasterizer returns soft radiopacity maps and exact semantic masks. The
compositor adds calcifications with local contrast and noise matching,
tissue-dependent visibility, point-spread smoothing, and a bounded local
processing halo. The same pipeline adds non-arterial microcalcification,
macrocalcification, ductal calcification, and metallic-object distractors as
explicit hard negatives. Candidate realizations are accepted only after
quality-control checks on generated area, length, component count, saturation,
and safe-region consistency; otherwise, the generator retries with a
deterministic seed offset and returns the best available candidate if all
attempts fail.
Generated images and dense targets were stored with PackBits compression to keep
the synthetic training set practical to store and decode during distributed
training.
\ifwithappendices%
Further implementation details are provided in
Appendix~\ref{app:synthetic-bac-generation-details}.
\fi
\begin{figure*}[t]
    \centering
    \includegraphics[width=\textwidth]{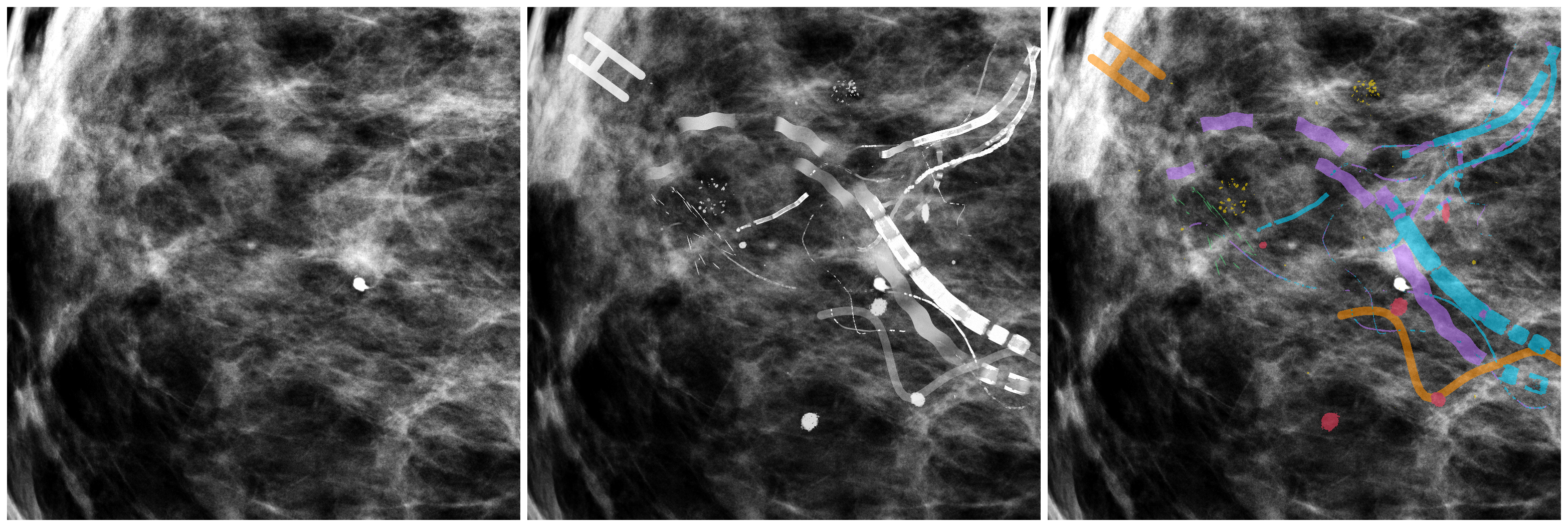}
    \caption{Synthetic preview generated from a de-identified mammogram crop.
    Panels show, from left to right, the source crop, the composited image with
    synthetic BAC, non-BAC calcifications, and metallic distractors, and the
    original crop with generated semantic masks overlaid. Cyan denotes regular
    BAC, yellow microcalcifications, pink macrocalcifications, green ductal
    calcifications, purple trace BAC, and orange metallic objects.}\label{fig:synthetic-bac-examples}
\end{figure*}

\subsection{Architecture}\label{subsec:architecture}

The BAC segmentation models use mammography-specific JEPA-pretrained Vision
Transformer (ViT) foundation encoders~\cite{Dosovitskiy2021,Assran2023}. ViT-S/16
and ViT-B/16 variants were pretrained on single-channel mammograms with a
joint-embedding predictive objective: a student observes masked context
regions, a teacher represents unmasked target regions, and a predictor matches
target embeddings in latent space rather than reconstructing pixels. This
initialization adapts image-token representations to mammographic anatomy and
acquisition statistics before BAC-specific supervision. A second predictor pass
also trained global \texttt{CLS} tokens, so the transferred encoders contain
both dense visual tokens and global summary tokens.

The ViT design follows the DINOv3 family in its use of SwiGLU feed-forward
blocks, two-dimensional rotary position encoding, register tokens, and
LayerScale~\cite{Simeoni2025DINOv3}. Both model sizes use 16~$\times$~16
patches, 12 transformer layers, 12 attention heads, four register tokens, and
four \texttt{CLS} tokens. The ViT-S/16 encoder uses hidden and feed-forward
dimensions of 384 and 1536, respectively, while ViT-B/16 uses 768 and 3072.

For BAC segmentation, we retained the pretrained backbone and replaced prior
heads with an image-resolution binary segmentation head. The backbone converts
each 2048~$\times$~1536 mammogram into a 128~$\times$~96 visual-token grid,
which is decoded through learned upsampling layers. A parallel high-resolution
pathway applies a stride-2 convolution and three ConvNeXt-style residual
blocks~\cite{Liu2022ConvNeXt} to the input mammogram, then fuses these features
into the decoder at half resolution. This combines transformer context with
local edge and texture cues for thin arterial calcifications. The final output
is a full-resolution BAC logit map.
\ifwithappendices%
Additional architectural details are provided in
Appendix~\ref{app:model-architecture-details}.
\fi

\subsection{Training}\label{subsec:training}

Supervised training used the synthetic mammograms and exact masks described
above. The task was binary BAC segmentation: regular and trace BAC were mapped
to foreground, while tissue, non-arterial calcifications, and metallic
distractors were mapped to background. Dense vessel-geometry maps were
generated for inspection and optional auxiliary training, but the experiments
reported here used a single BAC logit without an auxiliary geometry loss.

The segmentation objective combined foreground-weighted focal binary
cross-entropy~\cite{Lin2017FocalLoss} with soft Dice loss~\cite{Milletari2016VNet}.
The focal term used $\gamma=2.0$ and a positive target weight of 4.0, while the
Dice term was added with weight 0.1:
\begin{equation}
\mathcal{L}
= \mathcal{L}_{focal}(z,y;\gamma,w_+,m)
  + \lambda_D \mathcal{L}_{Dice}(z,y),
\label{eq:segmentation-loss}
\end{equation}
Here $z_i$ is the BAC logit, $y_i$ is the binary BAC target, $m_i$ is the
per-pixel mask weight, $w_+=4.0$, $\gamma=2.0$, and $\lambda_D=0.1$. We trained
with AdamW, separate peak learning rates of $5\times10^{-5}$ for the pretrained
backbone and $2\times10^{-4}$ for the newly initialized segmentation head, and a
one-cycle learning-rate schedule. The one-cycle schedule used a 5\% warmup
fraction, started at one-fifth of each peak learning rate, and ended at a factor
of 25 below the initial learning rate. All experiments were conducted on two
NVIDIA GeForce RTX 5090 GPUs with distributed data-parallel training, per-GPU
batch size 3, 16-step gradient accumulation, and an effective batch size of 96
images per optimizer step. Training ran for 45 epochs. Augmentations included
spatial cropping, rotation, flips, intensity perturbations, inversion, local
erasing, posterization, solarization, and Gaussian noise%
\ifwithappendices%
; exact settings are provided in
Appendix~\ref{app:bac-segmentation-training-details}%
\fi
.
Validation was performed after each epoch, and the selected checkpoint was the
one with the highest validation Dice.

\subsection{BacSeg Image-Level Evaluation}\label{subsec:bacseg-evaluation}

We evaluated image-level BAC classification on the full BacSeg dataset using
all three provided splits. Each image was labeled positive if its paired
ground-truth mask contained any nonzero BAC pixels; thresholds $m>0$ and
$m>127$ produced the same 877 positive and 123 negative labels. No BacSeg
images were used for training, checkpoint selection, or aggregation selection;
the split names refer only to the dataset provider's partitioning. We evaluated
at the image level because direct intersection-over-union (IoU) or Dice
comparison may penalize annotation-convention differences between pixel-perfect
synthetic targets and human-drawn masks rather than clinically relevant BAC
detection.

For each image, the grayscale PNG input was resized to the model input size
of 2048~$\times$~1536, scaled to $[0,1]$, normalized with the training
normalization, and passed through the trained segmentation model. The predicted
probability map was reduced to scalar image-level scores using the maximum
probability, mean probability, top-100 mean probability, top-1000 mean
probability, top 0.01\% mean probability, top 0.1\% mean probability, and
predicted positive area fraction at threshold $p \ge 0.5$. AUROC was computed
separately for each scalar score. We report all aggregations to make clear that
the best score is exploratory rather than prespecified.

\section{Results}
Table~\ref{tab:segmentation-results} summarizes held-out synthetic validation
performance. ViT-B achieved stronger overlap than ViT-S, increasing IoU from
0.5155 to 0.5325 and Dice from 0.6189 to 0.6357, but training time increased
from approximately 8~h to 13~h on two RTX 5090 GPUs. Qualitative review of
predicted masks indicated that the soft Dice term helped steer optimization
toward crisp, spatially coherent segmentation outputs.

\begin{table}[t]
    \centering
    \caption{Synthetic validation segmentation performance.\label{tab:segmentation-results}}
    \begin{tabular}{lccc}
        \toprule
        Model & IoU $\uparrow$ & Dice $\uparrow$ & Training time $\downarrow$ \\
        \midrule
        ViT-S & 0.5155 & 0.6189 & \textbf{8~h} \\
        ViT-B & \textbf{0.5325} & \textbf{0.6357} & 13~h \\
        \bottomrule
    \end{tabular}
\end{table}

Table~\ref{tab:bacseg-auroc} reports independent image-level BacSeg
classification across all predefined probability-map aggregations. Across all
1,000 images, ViT-B AUROC ranged from 0.7750 to 0.8719; the highest post hoc
score used top 0.01\% mean probability. Other sparse summaries were similar,
including top-100 mean probability (0.8664) and positive area fraction at
$p \ge 0.5$ (0.8606). The all-split result was lower than the test-only result:
the best ViT-S score dropped from 0.9049 to 0.8547, and the best ViT-B score
from 0.9226 to 0.8719.

\begin{table}[t]
    \centering
    \caption{All-split image-level BAC classification AUROC on BacSeg.\label{tab:bacseg-auroc}}
    \begin{tabular}{lcc}
        \toprule
        Aggregation score & ViT-S AUROC $\uparrow$ & ViT-B AUROC $\uparrow$ \\
        \midrule
        Max probability & 0.8143 & 0.7750 \\
        Mean probability & 0.8420 & 0.8567 \\
        Top-100 mean probability & 0.8476 & 0.8664 \\
        Top-1000 mean probability & 0.8511 & 0.8685 \\
        Top 0.01\% mean probability & \textbf{0.8547} & \textbf{0.8719} \\
        Top 0.1\% mean probability & 0.8459 & 0.8615 \\
        Area fraction, $p \ge 0.5$ & 0.8464 & 0.8606 \\
        \bottomrule
    \end{tabular}
\end{table}

We also tested auxiliary supervision from synthetic vessel-direction and
vessel-width targets. These targets gave negligible BAC segmentation improvement
and were omitted because the additional high-resolution arrays increased
input/output overhead without producing useful standalone outputs.

Table~\ref{tab:inference-benchmark} reports model sizes and compiled
forward-pass benchmarks for batch size 4, reflecting the four standard
mammography views in a screening examination. The benchmark used normalized
tensors with shape $(4,1,2048,1536)$ and excluded checkpoint loading,
preprocessing, sigmoid/export, TIFF writing, and overlay generation. ViT-S and
ViT-B contained 29.43M and 114.97M trainable parameters, respectively. On an RTX
5090, ViT-S processed a four-view batch in 110.68~ms and ViT-B in 213.63~ms. CPU
inference on an AMD EPYC 7763 was feasible but substantially slower.

\begin{table*}[t]
    \centering
    \caption{Compiled segmentation inference benchmark with batch size 4.\label{tab:inference-benchmark}}
    \begin{tabular}{llrrrrl}
        \toprule
        Model & Device & Params $\downarrow$ & Mean/batch $\downarrow$ &
        Mean/image $\downarrow$ & Throughput $\uparrow$ & Peak memory $\downarrow$ \\
        \midrule
        ViT-S & RTX 5090 & \textbf{29.43M} & \textbf{110.68~ms} &
        \textbf{27.67~ms} & \textbf{36.14 images/s} &
        \textbf{2.10~GiB allocated; 3.30~GiB reserved} \\
        ViT-B & RTX 5090 & 114.97M & 213.63~ms & 53.41~ms & 18.72 images/s &
        3.02~GiB allocated; 3.72~GiB reserved \\
        ViT-S & AMD EPYC 7763 & \textbf{29.43M} & \textbf{12.35~s} &
        \textbf{3.09~s} & \textbf{0.32 images/s} &
        \textbf{4.50~GiB high-water RSS} \\
        ViT-B & AMD EPYC 7763 & 114.97M & 20.76~s & 5.19~s & 0.19 images/s &
        5.90~GiB high-water RSS \\
        \bottomrule
    \end{tabular}
\end{table*}

Synthetic-image generation was also fast enough for iterative dataset
construction. The end-to-end Criterion benchmark generated one severe
2048~$\times$~1536 synthetic image per iteration. As shown in
Table~\ref{tab:synthesis-benchmark}, mean runtime was 2.7071~s per image
(95\% CI, 2.627--2.752~s) on an AMD EPYC 7763 system with 64 physical CPU cores.
This is a single-output benchmark because each iteration generates one
synthetic image. It is not strictly single-core: the generator uses Rayon
parallelism for selected large-image operations, and independent images can be
distributed across worker processes or batch jobs to increase aggregate
throughput.

\begin{table}[t]
    \centering
    \caption{End-to-end synthetic BAC generation benchmark.\label{tab:synthesis-benchmark}}
    \begin{tabular}{lr}
        \toprule
        Workload & Result \\
        \midrule
        Image size & 2048~$\times$~1536; 3,145,728 pixels \\
        Severity & Severe \\
        Mean runtime & 2.7071~s/image \\
        95\% CI & 2.6274--2.7518~s \\
        Median runtime & 2.7454~s/image \\
        Throughput & 1.1620~Melem/s \\
        Throughput CI & 1.1431--1.1973~Melem/s \\
        Samples; outliers & 10; 2/10 \\
        CPU & AMD EPYC 7763, 64 physical cores \\
        Software & Linux 6.8.0; rustc 1.92.0 \\
        \bottomrule
    \end{tabular}
\end{table}

\begin{figure*}[t]
    \centering
    \begin{minipage}[t]{0.48\textwidth}
        \centering
        \includegraphics[width=0.95\linewidth]{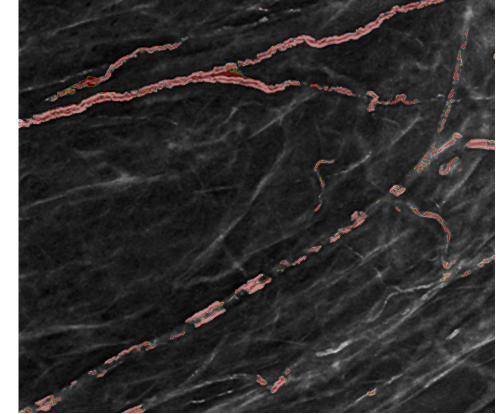}\\
        (a) Tight BAC segmentation.
    \end{minipage}\hfill
    \begin{minipage}[t]{0.48\textwidth}
        \centering
        \includegraphics[width=0.78\linewidth]{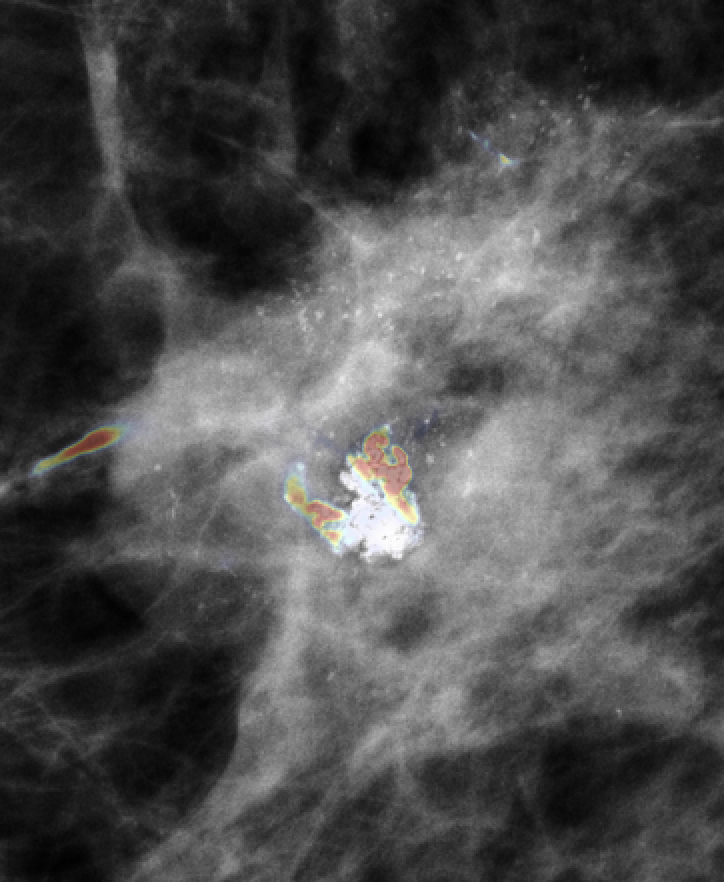}\\
        (b) False positive on non-arterial calcification.
    \end{minipage}
    \caption{Qualitative segmentation examples. The close-up example shows
    tight predicted BAC outlines enabled by pixel-perfect synthetic supervision
    and the high-resolution pathway fused into the ViT decoder. The false
    positive example shows residual activation on a non-arterial calcification;
    synthetic non-arterial calcification distractors reduce this failure mode
    but do not fully eliminate it in real mammograms.\label{fig:qualitative-examples}}
\end{figure*}

Figure~\ref{fig:qualitative-examples} illustrates qualitative outputs on real
mammograms. The high-resolution pathway appeared critical for producing tight
outlines around thin calcified tracks by preserving local edge and texture cues
that are difficult to recover from the ViT token grid alone.

\section{Discussion}
These results support the feasibility of training BAC segmentation models from
synthetic supervision when expert masks are unavailable. ViT-B provided the
best synthetic validation performance and strongest BacSeg image-level AUROC,
but its gains over ViT-S were modest relative to the added training and
inference cost. ViT-S may therefore be preferable for high-throughput screening
or CPU-constrained environments, while ViT-B may be justified when overlap or
image-level discrimination is the priority.

The synthetic and BacSeg evaluations answer different questions. Exact
synthetic masks support controlled training and debugging, but generated-data
overlap can overstate clinical generalization if the model learns
generator-specific appearance, boundary conventions, or distractor statistics.
BacSeg provides independent human-labeled evidence that the maps contain useful
BAC signal at the image level. However, image-level AUROC discards localization
and burden information and does not establish pixel-level agreement or
calibrated BAC area on real mammograms. The external validation should
therefore be interpreted as image-level transfer to human-labeled
synthetic 2D mammograms, not as completed clinical segmentation validation.

BAC-free background selection also introduces bias. We used an open-source BAC
segmentation model to screen for low-score backgrounds, so images with
false-positive response from metallic objects or breast implants may be
underrepresented in the training pool. This makes explicit metallic-object
distractors, implant-aware augmentation, and other high-contrast confounders
important for future synthetic data development. Breast implant false positives
can be mitigated by using DICOM metadata to identify implant studies and, when
available, prioritizing implant-displaced views for BAC assessment.

The analysis is limited to two-dimensional mammography, and qualitative review
suggested poor transfer to digital breast tomosynthesis (DBT). Individual DBT
slices were not uniformly poor, but maximum-intensity projections produced many
false positives, consistent with tomosynthesis reconstruction and projection
artifacts. Future DBT work should leverage accompanying synthetic 2D views,
either by mapping real BAC annotations from the S-view onto the DBT volume or by
using the 2D view to identify BAC-free DBT templates for synthetic injection.

Another high-value next step is clinician-in-the-loop annotation. The model can
be tuned for high sensitivity and used to pre-segment candidate calcifications
with crisp boundaries, after which expert clinicians can erase false positives
and refine true BAC annotations. This would use synthetic supervision to reduce
the burden of drawing thin arterial structures from scratch while producing
real expert-reviewed masks for validation and fine-tuning.

Although this study reports pixel-overlap metrics, the segmentation output can
also be converted to physical BAC burden using DICOM pixel spacing, providing a
path to downstream risk modeling, longitudinal comparison, and severity
stratification.

\section{Conclusion}
We presented a synthetic-supervision pipeline for BAC segmentation using
JEPA-pretrained mammography Vision Transformers. On held-out synthetic
validation data, ViT-B achieved the best overlap metrics with IoU 0.5325 and
Dice 0.6357, while ViT-S provided a lower-cost alternative with IoU 0.5155 and
Dice 0.6189. On the independent BacSeg dataset, the highest post hoc
image-level AUROC on all 1,000 available human-labeled synthetic 2D
mammograms was 0.8719 for ViT-B and 0.8547 for ViT-S. Runtime benchmarks show
that four-view GPU inference is practical for both model sizes and that
synthetic BAC generation is fast enough for iterative dataset construction on a
multicore workstation. The next essential steps are to improve synthetic realism and
confounder coverage, use the trained model to accelerate expert-reviewed
real-mammogram annotation, and extend the synthetic-supervision strategy to DBT
using accompanying synthetic 2D views.

\ifwithappendices%
\appendices%
\section{BAC-Free Background Selection Details}

For reproducibility, SCU-Net screening used the published 512~$\times$~512
model configuration from Guo et al.~\cite{Guo2021}. DICOM mammograms were
decoded as single-frame 2D images, normalized to 8-bit grayscale input,
median-filtered, and processed with tiled 512~$\times$~512 inference. The
ranking score for image $I$ was
\begin{equation}
    s(I) =
    \frac{\sum_{p \in I} \max(z_p, 0)}
         {\sum_{p \in I} \mathbb{1}[I_p > 0]},
\end{equation}
where $z_p$ is the raw SCU-Net output at pixel $p$ and the denominator is the
nonzero breast foreground area. The screening procedure successfully scored
283,465 images. Of these, 207,615 training images, 8,184 development images,
and 8,767 held-out test images had matched study-partition metadata; images
without matched partition metadata were excluded from background selection. All
10,000 selected training backgrounds had score 0.0; the selected development
backgrounds had maximum score
$9.91 \times 10^{-7}$.

\section{Synthetic BAC Generation Details}\label{app:synthetic-bac-generation-details}

The synthetic BAC generator is implemented as a Rust core with Python bindings.
The public image contract uses channel-last arrays with shape
$(F,H,W,C)$ for frames, height, width, and channels. The current generator
supports one-channel 16-bit mammograms and one-channel 8-bit semantic masks.
Mask label 0 denotes background, label 1 denotes BAC, label 2 denotes optional
microcalcifications, label 3 denotes optional macrocalcifications, label 4
denotes optional ductal calcifications, label 5 denotes trace BAC, and label 6
denotes metallic objects when the metallic layer is the most radiopaque
overlapping rendered feature. Metallic objects are also exposed through a
dedicated metallic mask because they are radiopaque distractors rather than
calcification classes.

All geometry is sampled in millimeters and converted to pixels only at
rasterization boundaries. The graph stage creates root-to-target vessel
centerlines, branches, local bends, and occasional projected loops while
preserving an overall arterial trajectory. Calcification sampling separates
burden from morphology: severity presets specify burden and radiopacity, while
deposit dimensions use a moderate morphology profile by default to avoid
confounding severity with unrealistically large vessel diameter. Supported
presets are trace, mild, moderate, marked, severe, and occlusive.

For each calcified vessel, the generator samples longitudinal disease runs and
deposits. Runs may be center-cut trace, vessel-constrained
microcalcification-like stretches, or mixed runs that alternate between those
two appearances. The previous solid full-diameter run style is not used.
Deposits are rendered as anisotropic granular particles with smoothed pores and
deterministic local width jitter. A faint vessel-background layer may also be
composited separately from the calcification mask, allowing vascular support to
be visible without being counted as BAC\@. A separate faint trace-BAC layer is
returned as semantic label 5 for subtle vessel-wall signal on calcified
vessels. In contrast to regular label-1 BAC, trace BAC is lower radiopacity,
does not define the primary calcified-burden measurement, and is kept separate
so experiments can either inspect it independently or merge it into foreground
BAC for binary segmentation.

The default realism profile samples radiopacity, deposit width, porosity, pore
scale, point-spread width, segment length, gap scale, and granule density from
a manually accepted set of candidate settings. Fixed seeds make each
realization deterministic, including sampled realism parameters.

Non-BAC calcification simulation is used to expose the segmentation model to
radiopaque findings that should not be labeled as BAC\@. Microcalcification
simulation creates up to three small pleiomorphic clusters per image, with
salt-like high-intensity particles that are not constrained to vessel walls; a
separate scattered path can add individual microcalcifications across the
breast support. Macrocalcification simulation creates larger, near-white,
porous, irregular calcifications whose count and brightness scale with the
selected severity preset but whose geometry remains non-arterial. Ductal
calcification simulation adds up to three fine-linear or fine-linear branching
groups in a linear or segmental distribution, with centers excluded from a
two-vessel-radius band around generated vessel centerlines. During binary BAC
training, semantic labels 2, 3, and 4 can therefore be mapped to background
while retaining their identity for inspection or multi-class experiments.

Metallic-object simulation targets a separate false-positive mode: general
high-radiopacity objects such as straight markers, smooth wires, long wires,
coiled wires, clips, rings, metal BBs, scar markers, and plate-like edge
artifacts. Generic metallic objects are sampled independently of the arterial
safe-insertion region so that they may appear inside, outside, or crossing
breast tissue, while skin-marker-like objects are constrained to breast tissue
support. They are composited as bright, low-porosity structures with sharper
edges and optional local artifact halos, and are returned through a dedicated
metallic mask and semantic label 6. This lets BAC models see metallic
radiopaque distractors during training without treating them as calcified
arterial burden.

For reproducibility, Fig.~\ref{fig:synthetic-bac-examples} was generated from
a de-identified mammogram crop using seed 6060, at crop rows 1900--2924 and
columns 1600--2624. The severe BAC preset was used with clustered and
scattered microcalcification, macrocalcification, ductal calcification, and
metallic-object sampling forced on. The visualization is a metadata-free PNG
assembled from the source crop, the composited image, and an overlay of the
generated semantic labels. The accompanying figure metadata records the
generator commit, crop, seed, configuration, label counts, and quality-control
summary.

\section{Model Architecture and Training Details}\label{app:model-architecture-details}

\subsection{JEPA Foundation Pre-Training}

The transferred encoder was initialized from a mammography JEPA foundation
model trained on the same general image domain before BAC-specific supervision.
Pretraining followed the I-JEPA principle of predicting latent target
representations from visible context representations~\cite{Assran2023}. For
each mammogram, contiguous non-overlapping context and target regions were
sampled over the patch-token grid. The student encoder processed the context
tokens, while an exponential-moving-average teacher encoded the corresponding
unmasked image targets. A cross-attention predictor, using target positional
embeddings as queries, predicted the teacher target embeddings from the student
context embeddings. The teacher was updated by exponential moving average after
optimizer steps, so the target representation evolved smoothly during training.

The pretraining design differed from standard I-JEPA in how global
\texttt{CLS} tokens were learned. The ViT backbone included four \texttt{CLS}
tokens in addition to visual patch tokens and register tokens. After the normal
visual-token JEPA predictor pass, a second predictor pass was performed using
only the student's \texttt{CLS} tokens as the context input. This second pass
used the same teacher target embeddings as the visual-context pass, forcing the
\texttt{CLS} tokens to carry image-level information that could predict masked
visual targets. A small regularization term was also applied to the student
\texttt{CLS} outputs to keep their representation well conditioned for
downstream probing. Separately, a Gram-style feature correlation loss was
introduced later in training to improve dense visual-token quality.

Pretraining combined the self-supervised JEPA losses with auxiliary supervised
probes for mammography tasks, including image-level triage, density, view,
implant detection, and lesion heatmap prediction when labels were available.
These probes encouraged clinically relevant global and spatial representations
but were discarded for BAC segmentation. The transferred ViT variants used here
were trained for 375 epochs, beginning at 256~$\times$~192 resolution and
increasing to 512~$\times$~384 for the final stage. Context and target masks
used fixed fractions of the token grid, with larger contiguous blocks at the
higher resolution stage to preserve comparable physical context.

\subsection{BAC Segmentation Architecture}

The BAC segmentation experiments used ViT-S/16 and ViT-B/16 backbone
configurations from the pretrained mammography JEPA foundation model. Both
encoders accept one-channel mammograms, use 16~$\times$~16 image patches, 12
transformer layers, 12 attention heads, four register tokens, four
\texttt{CLS} tokens, SwiGLU feed-forward blocks, LayerScale initialized to
$10^{-5}$, and two-dimensional rotary position encoding. ViT-S/16 uses hidden
size 384 and feed-forward size 1536; ViT-B/16 uses hidden size 768 and
feed-forward size 3072. BAC segmentation uses 2048~$\times$~1536 inputs,
yielding a 128~$\times$~96 visual-token grid before segmentation decoding. The
prior task-specific classification and detection heads are not used for BAC
segmentation; only the visual-token grid is passed to the segmentation decoder.

The segmentation decoder applies dropout probability 0.0 to the visual-token
grid and upsamples features through four resize-convolution stages with output
channels 128, 64, 32, and 1. Each stage is followed by two 3~$\times$~3
convolutions used as residual smoothing. The high-resolution image pathway
applies an initial 3~$\times$~3 stride-2 convolution from the source mammogram
to 32 channels, followed by GELU and three ConvNeXt-style residual blocks. Each
block uses a depthwise 7~$\times$~7 convolution, channel-wise LayerNorm, a
GELU-activated pointwise expansion by a factor of four, and residual addition.
These image-derived 32-channel features are fused into the decoder at the
1024~$\times$~768 stage, after which the decoder upsamples to a one-channel
full-resolution BAC logit map.

\subsection{BAC Segmentation Training}\label{app:bac-segmentation-training-details}

During supervised training, synthetic BAC semantic masks are converted to
binary targets by preserving labels 1 and 5 as foreground and mapping all other
labels to background. In the reported configurations, trace BAC is assigned a
foreground target of 1.0. The reduced-hybrid models use a single BAC logit and
disable auxiliary vessel-geometry loss; the generated dense vessel-geometry maps
remain available for optional experiments that add three geometry output
channels.

The loss is the sum of a foreground-weighted focal binary cross-entropy term and
a weighted soft Dice term. The focal term uses $\gamma=2.0$ and multiplies
foreground targets by a positive weight of 4.0; the Dice term is added with
weight 0.1. The optimizer is AdamW with weight decay 0.05, betas
$(0.85,0.95)$, and fused CUDA execution. The pretrained backbone uses peak
learning rate $5\times10^{-5}$, while the newly initialized segmentation head
uses peak learning rate $2\times10^{-4}$. Learning rates are scheduled with
OneCycleLR using \texttt{pct\_start}=0.05, \texttt{div\_factor}=5, and
\texttt{final\_div\_factor}=25. Training runs for 45 epochs with per-GPU batch
size 3, 16-step gradient accumulation, 16 data-loader workers per process, and
mixed-precision CUDA autocast. With two RTX 5090 GPUs, the effective batch size
is 96 images per optimizer step.

Training images and masks are randomly resized-cropped to 2048~$\times$~1536
with scale range 0.8--1.0 and aspect-ratio range 0.75--1.33, then rotated by
up to 10 degrees with probability 0.25 and independently flipped horizontally
and vertically with probability 0.5 each. Image-only augmentations include
posterization to 6 bits with probability 0.05, solarization at threshold 0.85
with probability 0.05, random percentile clipping with probability 0.5,
full-image inversion with probability 0.5, one to three local inversions with
probability 0.25, brightness and contrast jitter of 0.15, Gaussian noise with
standard deviation sampled from 0.005--0.03 with probability 0.5, and random
erasing with probability 0.2. Spatial augmentations are applied consistently to
images, masks, loss weights, and geometry targets where alignment is required.
Validation uses deterministic resize and the same normalization as training.
Validation reports loss, Dice, and intersection-over-union after each epoch,
and the selected model is the checkpoint with the highest validation Dice.

\fi

\section*{Acknowledgment}
The authors thank Tim Cogan for insightful discussions and MedCognetics, Inc.,
for data and compute resources.

\bibliographystyle{IEEEtran}
\bibliography{references}

@misc{CSBI2023,
  author = {{Canadian Society of Breast Imaging}},
  title = {{Canadian Society of Breast Imaging Position Statement on Breast Arterial Calcification Reporting on Mammography}},
  year = {2023},
  month = jan,
  url = {https://csbi.ca/canadian-society-of-breast-imaging-position-statement-on-breast-arterial-calcification-reporting-on-mammography/},
  note = {Accessed: 2026-05-25}
}

@article{Iribarren2022,
  author = {Iribarren, Carlos and Chandra, Malini and Lee, Catherine and Sanchez, Gabriela and Sam, Danny L. and Azamian, Farima Faith and Cho, Hyo-Min and Ding, Huanjun and Wong, Nathan D. and Molloi, Sabee},
  title = {Breast Arterial Calcification: A Novel Cardiovascular Risk Enhancer Among Postmenopausal Women},
  journal = {Circulation: Cardiovascular Imaging},
  volume = {15},
  number = {3},
  year = {2022},
  month = mar,
  doi = {10.1161/CIRCIMAGING.121.013526}
}

@article{Allen2024,
  author = {Allen, Tara Shrout and Bui, Quan M. and Petersen, Gregory M. and Mantey, Richard and Wang, Junhao and Nerlekar, Nitesh and Eghtedari, Mohammad and Daniels, Lori B.},
  title = {Automated Breast Arterial Calcification Score Is Associated With Cardiovascular Outcomes and Mortality},
  journal = {JACC: Advances},
  volume = {3},
  number = {11},
  pages = {101283},
  year = {2024},
  month = nov,
  doi = {10.1016/j.jacadv.2024.101283}
}

@article{Dapamede2026,
  author = {Dapamede, Theodorus and Urooj, Aisha and Joshi, Vedant and Gershon, Gabrielle and Li, Frank and Chavoshi, Mohammadreza and Brown-Mulry, Beatrice and Isaac, Rohan Satya and Mansuri, Aawez and Robichaux, Chad and Ayoub, Chadi and Arsanjani, Reza and Sperling, Laurence and Gichoya, Judy and van Assen, Marly and O'Neill, W. Charles and Banerjee, Imon and Trivedi, Hari},
  title = {Artificial Intelligence-Based Quantification of Breast Arterial Calcifications to Predict Cardiovascular Morbidity and Mortality},
  journal = {European Heart Journal},
  volume = {47},
  number = {18},
  pages = {2206--2220},
  year = {2026},
  month = mar,
  doi = {10.1093/eurheartj/ehag128}
}

@article{AlGhamdi2020,
  author = {AlGhamdi, Manal and Abdel-Mottaleb, Mohamed and Collado-Mesa, Fernando},
  title = {{DU-Net}: Convolutional Network for the Detection of Arterial Calcifications in Mammograms},
  journal = {IEEE Transactions on Medical Imaging},
  volume = {39},
  number = {10},
  pages = {3240--3249},
  year = {2020},
  month = oct,
  doi = {10.1109/TMI.2020.2989737}
}

@article{Guo2021,
  author = {Guo, Xiaoyuan and O'Neill, W. Charles and Vey, Brianna and Yang, Tianen Christopher and Kim, Thomas J. and Ghassemi, Maryzeh and Pan, Ian and Gichoya, Judy Wawira and Trivedi, Hari and Banerjee, Imon},
  title = {{SCU-Net}: A Deep Learning Method for Segmentation and Quantification of Breast Arterial Calcifications on Mammograms},
  journal = {Medical Physics},
  volume = {48},
  number = {10},
  pages = {5851--5861},
  year = {2021},
  month = aug,
  doi = {10.1002/mp.15017}
}

@article{Lee2017CBISDDSM,
  author = {Lee, Rebecca Sawyer and Gimenez, Francisco and Hoogi, Assaf and Miyake, Katelyn K. and Gorovoy, Mia and Rubin, Daniel L.},
  title = {A Curated Mammography Data Set for Use in Computer-Aided Detection and Diagnosis Research},
  journal = {Scientific Data},
  volume = {4},
  pages = {170177},
  year = {2017},
  month = dec,
  doi = {10.1038/sdata.2017.177}
}

@article{HallingBrown2021,
  author = {Halling-Brown, Mark D. and Warren, Lucy M. and Ward, Dominic and Lewis, Emma and Mackenzie, Alistair and Wallis, Matthew G. and Wilkinson, Louise S. and Given-Wilson, Rosalind M. and McAvinchey, Rita and Young, Kenneth C.},
  title = {{OPTIMAM} Mammography Image Database: A Large-Scale Resource of Mammography Images and Clinical Data},
  journal = {Radiology: Artificial Intelligence},
  volume = {3},
  number = {1},
  pages = {e200103},
  year = {2021},
  month = jan,
  doi = {10.1148/ryai.2020200103}
}

@inproceedings{Wang2022,
  author = {Wang, Kaier and Hill, Melissa and Knowles-Barley, Seymour and Tikhonov, Aristarkh and Litchfield, Lester and Bare, James Christopher},
  title = {Improving Segmentation of Breast Arterial Calcifications from Digital Mammography: Good Annotation Is All You Need},
  booktitle = {Proceedings of the Asian Conference on Computer Vision Workshops},
  pages = {130--146},
  year = {2022},
  month = dec
}

@article{Alamir2023,
  author = {Alamir, Manal and AlGhamdi, Manal and Collado-Mesa, Fernando and Abdel-Mottaleb, Mohamed},
  title = {Difference-of-{G}aussian Generative Adversarial Network for Segmenting Breast Arterial Calcifications in Mammograms},
  journal = {Expert Systems with Applications},
  volume = {217},
  pages = {119506},
  year = {2023},
  month = may,
  doi = {10.1016/j.eswa.2023.119506}
}

@article{AlJabri2024,
  author = {AlJabri, Manar and Alghamdi, Manal and Collado-Mesa, Fernando and Abdel-Mottaleb, Mohamed},
  title = {Recurrent Attention {U-Net} for Segmentation and Quantification of Breast Arterial Calcifications on Synthesized 2{D} Mammograms},
  journal = {PeerJ Computer Science},
  volume = {10},
  pages = {e2076},
  year = {2024},
  month = may,
  doi = {10.7717/peerj-cs.2076}
}

@article{Li2025,
  author = {Li, Wenbo and Zhang, Qiyu and Black, Dale and Ding, Huanjun and Iribarren, Carlos and Shojazadeh, Alireza and Molloi, Sabee},
  title = {Quantification of Breast Arterial Calcification in Mammograms Using a {UNet}-Based Deep Learning for Detecting Cardiovascular Disease},
  journal = {Academic Radiology},
  volume = {32},
  number = {9},
  pages = {5028--5038},
  year = {2025},
  month = sep,
  doi = {10.1016/j.acra.2025.05.036}
}

@article{Badano2018,
  author = {Badano, Aldo and Graff, Christian G. and Badal, Andreu and Sharma, Diksha and Zeng, Rongping and Samuelson, Frank W. and Glick, Stephen J. and Myers, Kyle J.},
  title = {Evaluation of Digital Breast Tomosynthesis as Replacement of Full-Field Digital Mammography Using an In Silico Imaging Trial},
  journal = {JAMA Network Open},
  volume = {1},
  number = {7},
  pages = {e185474},
  year = {2018},
  month = nov,
  doi = {10.1001/jamanetworkopen.2018.5474}
}

@article{Sharma2019,
  author = {Sharma, Diksha and Graff, Christian G. and Badal, Andreu and Zeng, Rongping and Sawant, Purva and Sengupta, Aunnasha and Dahal, Eshan and Badano, Aldo},
  title = {Technical Note: In Silico Imaging Tools from the {VICTRE} Clinical Trial},
  journal = {Medical Physics},
  volume = {46},
  number = {9},
  pages = {3924--3928},
  year = {2019},
  month = jul,
  doi = {10.1002/mp.13674}
}

@inproceedings{Sizikova2023,
  author = {Sizikova, Elena and Saharkhiz, Niloufar and Sharma, Diksha and Lago, Miguel and Sahiner, Berkman and Delfino, Jana and Badano, Aldo},
  title = {Knowledge-Based In Silico Models and Dataset for the Comparative Evaluation of Mammography {AI} for a Range of Breast Characteristics, Lesion Conspicuities and Doses},
  booktitle = {Advances in Neural Information Processing Systems},
  volume = {36},
  year = {2023}
}

@article{Oyelade2022,
  author = {Oyelade, Olaide N. and Ezugwu, Absalom E. and Almutairi, Mubarak S. and Saha, Apu Kumar and Abualigah, Laith and Chiroma, Haruna},
  title = {A Generative Adversarial Network for Synthetization of Regions of Interest Based on Digital Mammograms},
  journal = {Scientific Reports},
  volume = {12},
  number = {1},
  year = {2022},
  month = apr,
  doi = {10.1038/s41598-022-09929-9}
}

@article{MontoyaDelAngel2024,
  author = {Montoya-del-Angel, Ricardo and Sam-Millan, Karla and Vilanova, Joan C. and Marti, Robert},
  title = {{MAM-E}: Mammographic Synthetic Image Generation with Diffusion Models},
  journal = {Sensors},
  volume = {24},
  number = {7},
  pages = {2076},
  year = {2024},
  month = mar,
  doi = {10.3390/s24072076}
}

@misc{Cho2026,
  author = {Cho, Hyunwoo and Kwon, Yeeun and Kim, Min Jung and Yoo, Yangmo},
  title = {{MC-GenRef}: Annotation-Free Mammography Microcalcification Segmentation with Generative Posterior Refinement},
  year = {2026},
  eprint = {2604.04470},
  archivePrefix = {arXiv},
  primaryClass = {eess.IV}
}

@inproceedings{Dosovitskiy2021,
  author = {Dosovitskiy, Alexey and Beyer, Lucas and Kolesnikov, Alexander and Weissenborn, Dirk and Zhai, Xiaohua and Unterthiner, Thomas and Dehghani, Mostafa and Minderer, Matthias and Heigold, Georg and Gelly, Sylvain and Uszkoreit, Jakob and Houlsby, Neil},
  title = {An Image Is Worth 16x16 Words: Transformers for Image Recognition at Scale},
  booktitle = {International Conference on Learning Representations},
  year = {2021}
}

@inproceedings{Assran2023,
  author = {Assran, Mahmoud and Duval, Quentin and Misra, Ishan and Bojanowski, Piotr and Vincent, Pascal and Rabbat, Michael and LeCun, Yann and Ballas, Nicolas},
  title = {Self-Supervised Learning from Images with a Joint-Embedding Predictive Architecture},
  booktitle = {Proceedings of the IEEE/CVF Conference on Computer Vision and Pattern Recognition},
  pages = {15619--15629},
  year = {2023}
}

@misc{Simeoni2025DINOv3,
  author = {Simeoni, Oriane and Vo, Huy V. and Seitzer, Maximilian and Baldassarre, Federico and Oquab, Maxime and Jose, Cijo and Khalidov, Vasil and Szafraniec, Marc and Yi, Seungeun and Ramamonjisoa, Michael and Massa, Francisco and Haziza, Daniel and Wehrstedt, Luca and Wang, Jianyuan and Darcet, Timothee and Moutakanni, Theo and Sentana, Leonel and Roberts, Claire and Vedaldi, Andrea and Tolan, Jamie and Brandt, John and Couprie, Camille and Mairal, Julien and Jegou, Herve and Labatut, Patrick and Bojanowski, Piotr},
  title = {{DINOv3}},
  year = {2025},
  eprint = {2508.10104},
  archivePrefix = {arXiv},
  primaryClass = {cs.CV}
}

@inproceedings{Liu2022ConvNeXt,
  author = {Liu, Zhuang and Mao, Hanzi and Wu, Chao-Yuan and Feichtenhofer, Christoph and Darrell, Trevor and Xie, Saining},
  title = {A ConvNet for the 2020s},
  booktitle = {Proceedings of the IEEE/CVF Conference on Computer Vision and Pattern Recognition},
  pages = {11976--11986},
  year = {2022}
}

@inproceedings{Lin2017FocalLoss,
  author = {Lin, Tsung-Yi and Goyal, Priya and Girshick, Ross and He, Kaiming and Dollar, Piotr},
  title = {Focal Loss for Dense Object Detection},
  booktitle = {Proceedings of the IEEE International Conference on Computer Vision},
  pages = {2980--2988},
  year = {2017}
}

@inproceedings{Milletari2016VNet,
  author = {Milletari, Fausto and Navab, Nassir and Ahmadi, Seyed-Ahmad},
  title = {{V-Net}: Fully Convolutional Neural Networks for Volumetric Medical Image Segmentation},
  booktitle = {Fourth International Conference on 3D Vision},
  pages = {565--571},
  year = {2016},
  doi = {10.1109/3DV.2016.79}
}

\end{document}